\documentclass{egpubl}
\usepackage{eg2021}

\ConferencePaper        %

\usepackage[T1]{fontenc}
\usepackage{dfadobe}
\usepackage{amsfonts}
\usepackage{amsmath}
\usepackage{xcolor}

\usepackage{cite}  %
\biberVersion
\BibtexOrBiblatex
\electronicVersion
\PrintedOrElectronic

\ifpdf \usepackage[pdftex]{graphicx} \pdfcompresslevel=9
\else \usepackage[dvips]{graphicx} \fi

\def\fg#1{Fig.~\ref{fig:#1}}

\def\perm{, used with permission}

\usepackage{amsmath}
\usepackage{egweblnk}
\usepackage{import}

\newif\iffinal

\finaltrue

\iffinal
\def\revision#1{#1}
\else
\def\revision#1{{\color{red}{#1}}}
\fi

\title[STALP]{STALP: Style Transfer with Auxiliary Limited Pairing}

\iffinal
\author[Futschik et al.]
{\parbox{\textwidth}{\centering D. Futschik$^{1}$, M. Ku\v{c}era$^{1}$, M. Luk\'{a}\v{c}$^{2}$, Z. Wang$^{2}$, E. Shechtman$^{2}$, D. S\'{y}kora$^{1}$ }\\
	{\parbox{\textwidth}{\centering
			$^1$Czech Technical University in Prague, Faculty of Electrical Engineering, Czech Republic\\
			$^2$Adobe Research, USA
		}
	}
}
\else
\author[paper1116]{\parbox{\textwidth}{\centering paper1116}}
\fi

\begin{document}

\teaser{\def\svgwidth{\hsize}
\begingroup%
  \makeatletter%
  \providecommand\color[2][]{%
    \errmessage{(Inkscape) Color is used for the text in Inkscape, but the package 'color.sty' is not loaded}%
    \renewcommand\color[2][]{}%
  }%
  \providecommand\transparent[1]{%
    \errmessage{(Inkscape) Transparency is used (non-zero) for the text in Inkscape, but the package 'transparent.sty' is not loaded}%
    \renewcommand\transparent[1]{}%
  }%
  \providecommand\rotatebox[2]{#2}%
  \ifx\svgwidth\undefined%
    \setlength{\unitlength}{1199.60927734bp}%
    \ifx\svgscale\undefined%
      \relax%
    \else%
      \setlength{\unitlength}{\unitlength * \real{\svgscale}}%
    \fi%
  \else%
    \setlength{\unitlength}{\svgwidth}%
  \fi%
  \global\let\svgwidth\undefined%
  \global\let\svgscale\undefined%
  \makeatother%
  \begin{picture}(1,0.173712)%
    \put(0,0){\includegraphics[width=\unitlength]{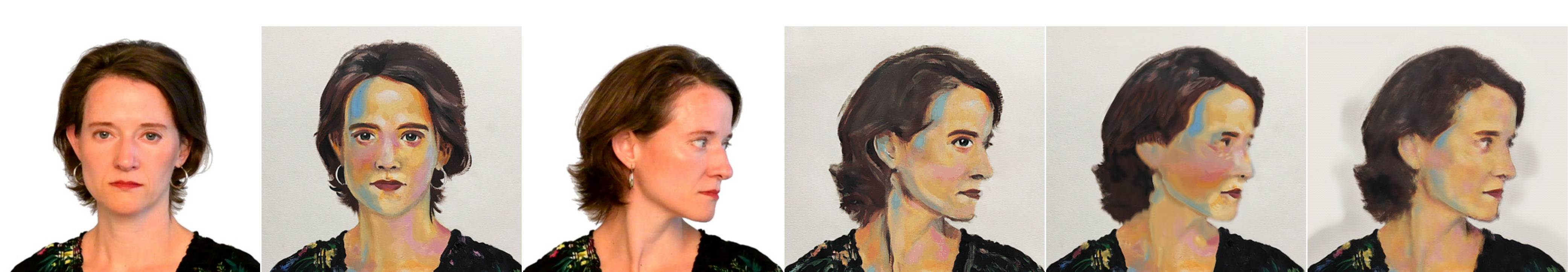}}%
    \put(0.24981355,0.162){\color[rgb]{0,0,0}\makebox(0,0)[b]{\smash{source style}}}%
    \put(0.08297481,0.162){\color[rgb]{0,0,0}\makebox(0,0)[b]{\smash{source frame}}}%
    \put(0.75132908,0.162){\color[rgb]{0,0,0}\makebox(0,0)[b]{\smash{Jamri\v{s}ka et al.}}}%
    \put(0.41689225,0.162){\color[rgb]{0,0,0}\makebox(0,0)[b]{\smash{target frame}}}%
    \put(0.58358054,0.162){\color[rgb]{0,0,0}\makebox(0,0)[b]{\smash{our approach}}}%
    \put(0.91750160,0.162){\color[rgb]{0,0,0}\makebox(0,0)[b]{\smash{Texler et al.}}}%
  \end{picture}%
\endgroup%
\caption{\revision{An example of style transfer
with auxiliary pairing---an artist prepares a stylized version (source style)
of a selected frame from input video (source frame). Then a network is trained
to transfer artist's style to remaining video frames (target frame). During the
training phase a subset of input video frames as well as the source frame and
its stylized counterpart are taken into account. Once the network is trained,
the entire sequence can be stylized in real-time (our approach). In contrast
to current state-of-the-art in example-based video stylzation (Jamri\v{s}ka et
al.~\cite{Jamriska19} and Texler et al.~\cite{Texler20b}) our approach better
preserves important visual characteristics of the style exemplar even though
the scene structure changed considerably (head rotation). \iffinal Input video
frames and source style~\copyright~Zuzana Studen\'{a}\perm.\fi}}\label{fig:zuzka2}\label{fig:teaser}}

\maketitle
\begin{abstract}

We present an approach to example-based stylization of images that uses a
single pair of a source image and its stylized counterpart. We demonstrate how
to train an image translation network that can perform real-time semantically
meaningful style transfer to a set of target images with similar content as the
source image. A key added value of our approach is that it considers also
consistency of target images during training. Although those have no stylized
counterparts, we constrain the translation to keep the statistics of neural
responses compatible with those extracted from the stylized source. In contrast
to concurrent techniques that use a similar input, our approach better preserves
important visual characteristics of the source style and can deliver temporally
stable results without the need to explicitly handle temporal consistency. We
demonstrate its practical utility on various applications including video
stylization, style transfer to panoramas, faces, and 3D models.

\ccsdesc[500]{Computing methodologies~Non-photorealistic rendering}
\printccsdesc
\end{abstract}

\section{Introduction}

In recent years, methods for performing automatic style transfer from an
exemplar image to a target image or a video have gained significant popularity.
Although state of the art in this field progresses quickly and produces ever
more believable artistic images, there are still aspects in which most methods
tend to have fundamental shortcomings. One such crucial element is defining the
semantic intent while still preserving visual characteristics of the used
artistic media.

A seminal work in this direction was the Image Analogies framework introduced
by Hertzmann et al.~\cite{Hertzmann01}, which requires the user to provide a
set of guidance channels~\cite{Benard13, Fiser16, Fiser17, Jamriska19} that
encourage the synthesis algorithm to transfer smaller patches of the style
exemplar onto desired spatial locations in the target image. Those channels,
however, need to be prepared explicitly by the user or generated
algorithmically for a limited target domain, e.g., 3D renders~\cite{Fiser16},
facial images~\cite{Fiser17}, or a sequence of video frames close to the
stylized keyframe~\cite{Jamriska19}. Deriving consistent semantically
meaningful guidance in the general case remains an open problem.

Neural approaches to style transfer~\cite{Gatys16, Li17, Kolkin19} rely on the
assumption that one can encode semantic similarity using the correspondence of
statistics of neural features extracted from responses of the VGG
network~\cite{Simonyan14}. Although such an assumption holds in some cases, it
is not easy to amend when it fails. Moreover, in contrast to patch-based
methods, neural techniques tend to produce noticeable visual artifacts due to
their statistical nature. One can partially alleviate this drawback by applying
patch-based synthesis in the neural domain~\cite{Li16b, Liao17}. However, since
in this scenario neural features are transferred explicitly, the requirement of
knowledge of accurate correspondences is still inevitable.

Another possibility of preserving semantically meaningful transfer is using the
image-to-image translation principle pioneered by Isola et al.~\cite{Isola17}.
This approach can encode semantic intent and retain high-quality output.
However, it has a fundamental limitation of requiring a relatively large
dataset of image pairs (original image plus its stylized counterpart), which is
rarely easy to obtain when considering artistic applications. Lastly, a group
of unpaired image translation algorithms could be used~\cite{Zhu17a, Park20},
however, since it can be difficult to incorporate intent into these methods,
they are not as suitable for tasks where the artist needs greater control.

In this paper, we present a novel approach to neural style transfer that allows
artists to stylize a set of images with arbitrary yet similar content in a
semantically meaningful way, while preserving the target subjects' critical
structural features. In contrast to previous neural techniques, in our
framework, the user explicitly encodes the semantic intent by specifying a
stylized counterpart for a selected image from the set that needs to be
stylized. Using this single style exemplar, we then train an image-to-image
translation network that stylizes the remaining images. Our approach bears a
resemblance to the recent keyframe-based video stylization framework of Texler
et al.~\cite{Texler20b}, where a similar workflow is used. A key difference in
our technique is that we consider other frames from the input sequence during
the training phase. This enables us to ensure temporal stability without
explicit guidance and better preserve style when the remaining video frames
deviate from the original keyframe. Moreover, thanks to this increased
robustness, our framework goes beyond video stylization. One can use it also in
more challenging scenarios, including auto-completion of a panorama painting,
stylization of 3D renders, or different portraits captured under similar
illumination conditions.

\section{Related Work}

Despite the renewed interest and broader impact, image stylization algorithms
date back decades. Traditionally, they were based on predefined, hand-designed
transformations limited to a subset of styles, and possibly target domains as
well. One example of such transformation approach was shown by Curtis et
al.~\cite{Curtis97}, running a physical simulation to produce watercolor filter
effect. Other research directions focused on composing images from static or
procedurally generated brush strokes or pens~\cite{Benard10, Bousseau06,
Praun01, Salisbury97}. These conventional algorithmic approaches can create
very appealing results, but they have the added difficulty of requiring the
style filters to be designed on an individual basis. Therefore, the act of
creating a new style or even slight modifications of existing styles tends to
necessitate considerable amounts of effort. These methods do not require a
style exemplar, but instead contain a prior given by the design of the filter.

The framework of Image Analogies proposed by Hertzmann et
al.~\cite{Hertzmann01} trades designing elements of the output image directly
for designing a set of guidance channels which form a loss function. Optimizing
over pixel locations and directly copying patches of an exemplar image
guarantees that features found in the exemplar will be represented exactly in
the resulting image. This framework became the basis of numerous style transfer
methods~\cite{Benard13, Fiser16, Fiser17, Dvoroznak18, Jamriska19}. A key
advantage over traditional algorithmic methods lies the fact that this
framework allows for transfer of arbitrary style.

However, creating the guidance channels is cumbersome, and in some potential
applications it might not be always clear how to design algorithms for
obtaining them automatically, and still, the task of preparing a framework that
would work with arbitrary images remains seemingly impossible. To sidestep this
issue, methods of general style transfer have been formulated. Frigo et
al.~\cite{Frigo16} attempts to re-imagine the problem of guiding channels by
splitting the image into partitions and matching these to their counterparts.
More commonly known, Gatys et al.~\cite{Gatys16} uses responses of a neural
network to generate global style statistics which an optimization process sees
to reproduce in the result while incorporating a content constraint to prevent
the overall structure from diverging too far from the target image.
\revision{Refining these ideas to a video domain and employing a more
sophisticated loss functions, others~\cite{Chen17, Li17, Ruder18, Kolkin19}
manage to produce results which are coherent in time and more faithful to the
style.} While they produce impressive results on some inputs, these methods
generally take all the control out of the artists' hands and are notoriously
difficult to steer in different directions, as their mechanisms are
non-intuitive and unpredictable.

A different view of the problem is offered by the image-to-image framework,
which aims to translate images from one domain to another, which is directly
applicable to style transfer. While the original image translation
methods~\cite{Isola17, Johnson16} require relatively large dataset to work
reliably, by their combination with generative adversarial
models~\cite{Goodfellow14, Zhu17a}, this requirement can be relaxed. Unlike
techniques based on image analogies, these methods tend to require substantial
amount of model training. And although patch-based synthesis~\cite{Fiser17} can
be used to generate a large number of image pairs on which one can train the
image-to-image translation network~\cite{Futschik19}, the problem of having
meaningful guidance remains.

Few-shot learning techniques~\cite{Liu19, Wang19}, as well as approaches based
on deformation transfer~\cite{Siarohin19, Siarohin19b} require only a single
style exemplar. However, they still need pre-training on large dataset of
specific target domains and thus are not applicable in general case. Moreover,
these techniques capture only the target subject's coarse deformation
characteristics; its structure or identity is omitted. A similar drawback also
holds for approaches based on generative adversarial networks such as StyleGAN
v2~\cite{Karras20}. In this approach, a massive collection of artworks is used
to train a network that can generate an artistic image for a given input latent
vector. Those vectors can then be predicted and fine-tuned to align the
generated image with the target image's features. However, this process is
inaccurate, leading to imprecise alignment that hinders the network's ability
to preserve the target subject's structure or identity.

\section{Our Approach} \label{sec:method}

\def\F{\mathcal{F}}

As input to our method, we take pairs of images~$K=(X,Y)$ called
\emph{keyframes}. They represent a visual translation from a source visual
domain of~$X$ into a target domain of~$Y$. For instance~$X$ can be a photo
and~$Y$ its stylized counterpart prepared by an artist (see~\fg{ours}). Note
that our key assumption about~$K$ is that it should be as small as possible, in
practice even a \emph{single} keyframe is usually sufficient. This is in line
with our central motivation to reduce the amount of manual work since the
creation of keyframes is time consuming and thus prohibitive. In addition
to~$K$, the user also provides a set of unpaired images~$Z$, which they would
like to stylize. The images in~$Z$ can be arbitrary, but our method works best
if their domain is similar or same as~$X$. For instance~$Z$ and~$X$ can be
frames from the same video sequence or photos from the same location, etc. If
there is a larger number of images in~$Z$, it is beneficial to prune it as
smaller number of images in~$Z$ usually has a positive effect on the resulting
quality (see~\fg{frames}). Both keyframes~$K$ as well as unpaired images~$Z$
are used during an optimization process that produces a neural translation
model~$\F$. Using~$\F$ one can stylize~$Z$ in a semantically meaningful way,
i.e., produce a set of output images~$O$ in which important visual features of
artistic style~$Y$ are reproduced at appropriate locations.

\begin{figure}[hb]
\def\svgwidth{\hsize}\import{figs/ours/}{ours.tex}\caption{An overview of our approach---we optimize weights~$\theta$ of
a translation network~$\F$ which accepts images from a source
domain~\revision{$X$ or~$Z$} and produces output images~\revision{$O$} with a
similar appearance as those in the target domain~$Y$. The high-frequency
details are preserved well, thanks to the~$L_1$ loss computed on the
artist-created style images~$Y$ which have the same structure as the input
images~$X$, while the style consistency on other images~$Z$ is enforced due
to the VGG loss. \iffinal\revision{Source style~\copyright~Graciela
Bombalova-Bogra\perm.}\fi}\label{fig:ours}
\medskip
\def\svgwidth{\hsize}\import{figs/netarch/}{netarch_s.tex}\caption{\revision{A network architecture used for
our model~$\F$: input layer~(green), one~$7\times7$ and two~$3\times3$
convolution blocks~(blue), nine~$3\times3$ residual blocks (yellow),
two~$3\times3$ upsampling blocks~(red), and one additional block
with~$7\times7$ convolutions~(blue). Skip connections~(black) are used to
connect downsampling and upsampling layers.}}\label{fig:netarch}
\end{figure}

As~$\F$, we use the network architecture design of Futschik et
al.~\cite{Futschik19} \revision{(see~\fg{netarch})}, a U-Net-type network,
which is particularly suitable for style transfer tasks as it allows to
reproduce important high-frequency details that are crucial for generating
believable artistic styles. In the original method of Futschik et al.~$\F$ was
trained on a large dataset of~$K$ which is intractable in our scenario. Texler
et al.~\cite{Texler20b} uses the network architecture of~$\F$ as well in a
similar setting as ours, i.e., small number of keyframes~$K$, however, their
method struggles with larger structural changes in the target images~$Z$.

To address this issue, we leverage the fact that the set of target images~$Z$
is known beforehand and thus we can incorporate this additional knowledge into
the optimization process. To do that, we introduce a different training
strategy. The process is a combination of two complementary objectives,
illustrated in~\fg{ours}, which we minimize as we train~$\F$:
\begin{itemize}
\item \emph{$L_1$ loss} on the original translation pairs~$K$, ensuring that
keyframes are represented as closely as possible.
\item \revision{\emph{VGG loss} between the images from set~$Z$ and set~$Y$,
which acts as a regularizer for the stylized images~$O$.}
\end{itemize}

Combining these two, we obtain the objective function we would like to minimize:
\begin{equation}\label{eqn:amtheta}
\sum_i | \F(X_i; \theta) - Y_i | + \lambda \sum_{j,k}{\sum_{l}{\lVert \mathcal{G}^l(\F(Z_j; \theta)) - \mathcal{G}^l(Y_k) \rVert^{2}}}
\end{equation}
where~$\theta$ is a set of weights of~$\F$ which we would like to optimize,
$\mathcal{G}^l$ stands for Gram correlation matrix calculated at layer~$l\in L$
after extracting VGG network responses~\cite{Simonyan14} of the given image,
and~$\lambda$ is a weighting coefficient which we set to~$100/(|Z||L|)$ for all
conducted experiments.

Contrary to previous techniques~\cite{Gatys16, Johnson16} which compute Gram
matrix from a subset of layers we found that evaluating the loss at every
layer~$l\in L$ of VGG is beneficial in terms of measuring the overall style
quality. However, this is computationally more expensive and thus our method
generally requires an order of magnitude more time to produce the final
results. These previous methods use the term purely as a proxy for style
transfer. In our case we use it as regularizer to prevent the model from
overfitting to the keyframes. This effect is visible in~\fg{ablation}, where
if we take away the VGG loss, the resulting~$\F$ is unable to generalize
beyond~$K$ whereas using VGG loss only will negatively affect the content.

\begin{figure*}[ht]
\def\svgwidth{\hsize}\import{figs/ablation/}{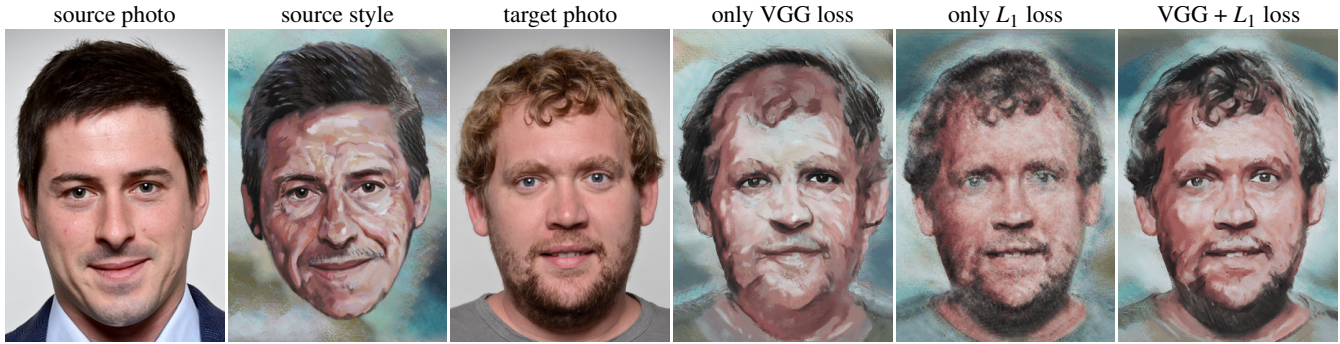}\caption{An ablation study demonstrating the importance of
individual terms in our objective function~(\ref{eqn:amtheta})---a stylized
pair~$(X_1, Y_1)$ (source photo, source style) is used together with~$Z_1$
(target photo) to optimize weights of model~$\F$. When only VGG loss is used,
the identity of a person in the target photo deteriorates. On the other hand
when only~$L_1$~loss is used during optimization source, style is not preserved
well. By combining~$L_1$~loss and VGG loss in~(\ref{eqn:amtheta}) we get the
result which produces a good balance between identity and style preservation.
\iffinal\revision{Source style~\copyright~Graciela Bombalova-Bogra\perm.}\fi}\label{fig:ablation}
\end{figure*}

By minimizing the objective~(\ref{eqn:amtheta}) we produce a trained
model~\revision{$\F$}, which in turn is able to stylize the images from~$Z$ via
a feed-forward pass. An important aspect to notice is that unlike most previous
style transfer techniques, our approach does not enforce any content loss
explicitly. We find that content losses found in literature~\cite{Gatys16,
Kolkin19} tend to be detrimental to the quality of style transfer, especially
when higher frequencies are concerned. It causes a particular washed-out look
where important style details are missing (see~\fg{content}). An objection to
our argument could be that without explicit penalty on the content preservation,
 the model can resort to memorizing the keyframes and return~$Y$ regardless the
content in target images~$Z$. This would eventually minimize both the~$L_1$
error as well as the VGG loss. The reason why the optimization process does not
end up using this trivial solution is twofold. We argue that due to the limited
receptive field of~$\F$, it has to learn an effective encoding of the input; in
addition, since the VGG loss is relatively weak and serves only as a non-linear
regularizer, it makes the trivial solution difficult to find during the
optimization process. Moreover, by optimizing a one-to-one mapping between
images of perceptually similar semantic structure ($X$ to~$Y$), we posit that
this acts as an implicit content preservation technique.

\begin{figure}[h]
\def\svgwidth{\hsize}\import{figs/content/}{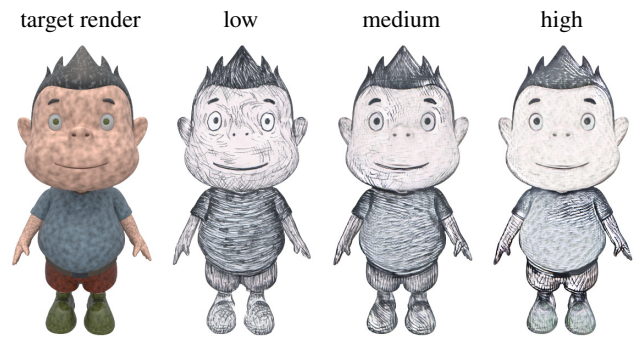}\caption{An illustration of a wash-out effect caused by adding an
explicit content loss term~\cite{Kolkin19} into our objective
function~(\ref{eqn:amtheta}). Target render stylized using model~$\F$ optimized
on a stylized pair from~\fg{boy} with low, medium, and high content loss
weight. Note how style details deteriorate gradually with the increasing
content loss. \iffinal\revision{Source style~\copyright~\v{S}t\v{e}p\'{a}nka
S\'{y}korov\'{a}\perm.}\fi}\label{fig:content}
\end{figure}

\section{Results}

We implemented our approach using PyTorch~\cite{Paszke19}. For all experiments,
we use Adam optimizer with learning rate~$10^{-4}$, $\beta_1=0.9$,
$\beta_2=0.999$. We found that higher rate does not work well when performing
many Gram matrix operations that are prone to producing exploding gradients.
For the network model~$\F$, we use~$9$ residual blocks, which is in line with
previous approaches~\cite{Futschik19, Texler20b}. However, since in our
optimization batch size is equal to~$1$ we use instance
normalization~\cite{Ulyanov16a} instead of batch normalization. All layers used
for Gram matrix computation are post-activated with ReLU to better incorporate
non-linearity. In each experiment, we let the optimization process run for
approximately 100k iterations, which translates into roughly 3--6 hours of wall
time on a single NVIDIA V100 GPU, depending on the target resolution. The
resolutions we produce range from 512px to 768px as longer side of the image,
with the shorter side scaled appropriately to preserve correct aspect ratio
given by the input images.

We evaluated our approach in five different use cases to demonstrate its wider
range of applicability: (1)~keyframe-based video stylization, (2)~style
transfer to 3D models, (3)~autopainting panorama images, (4)~example-based
stylization of portraits, and (5)~real-time stylization of video calls.

\begin{figure*}[ht]
\def\svgwidth{\hsize}\import{figs/video/}{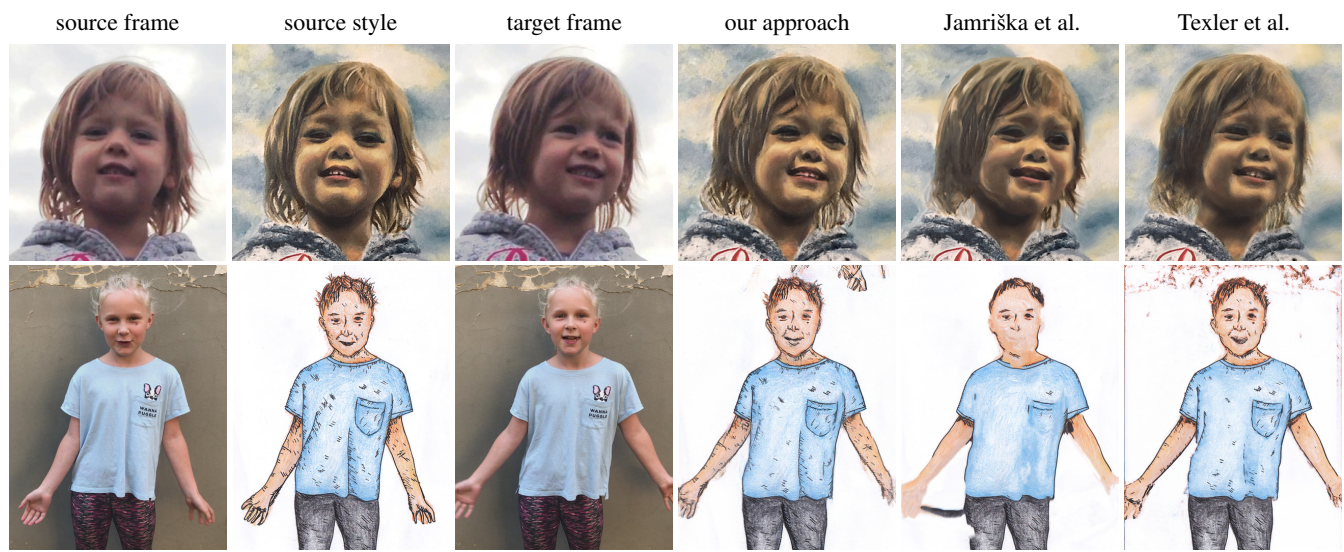}\caption{\revision{Video stylization results---in each
video sequence~(rows) a selected frame~(source frame) is stylized using
different artistic media~(source style). The network is then trained using this
stylized pair and a subset of frames from the entire video sequence~(target
frame). The results of our method (our approach) are compared with the output
of concurrent techniques: Jamri\v{s}ka et al.~\cite{Jamriska19} and Texler et
al.~\cite{Texler20b}. Note how our method better preserves important style
details and visual features of the target frames. Previous style transfer
techniques tend to produce wash out artifacts due to significant structural
changes with respect to the source frame. \iffinal Video frames and style~(top
row) \copyright~Zuzana Studen\'{a}, and~(bottom row)
\copyright~\v{S}t\v{e}p\'{a}nka S\'{y}korov\'{a}\perm.\fi}}\label{fig:video}
\end{figure*}

\begin{figure*}[ht]
\def\svgwidth{\hsize}\import{figs/mucha/}{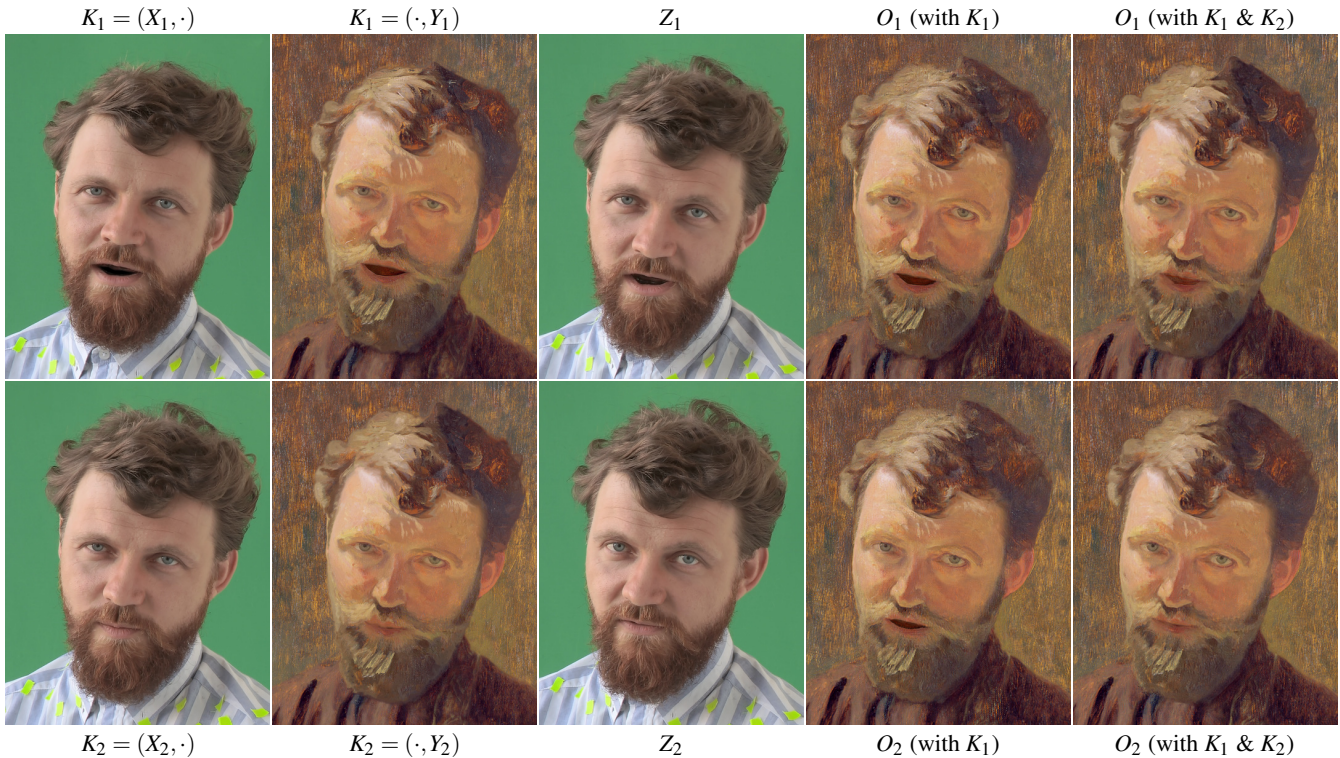}\caption{\revision{Example of video stylization with multiple
keyframes---two keyframes~$K_1=(X_1, Y_1)$ and~$K_2=(X_2, Y_2)$ were created by
painting over the input video frames~$X_1$ \&~$X_2$ to get their stylized
counterparts~$Y_1$ \&~$Y_2$. First, our network~$\F$ was trained using only
single keyframe~$K_1$ and applied to stylize input video frames~$Z_1$ \&~$Z_2$
to produce~$O_1$ \&~$O_2$ (with~$K_1$). Note, how closed mouth in~$Z_2$ was not
stylized properly in~$O_2$ (with~$K_1$). By adding $K_2$ to the list of
keyframes used during training phase, open and closed mouth is stylized better,
see~$O_1$ \&~$O_2$ (with~$K_1$ \&~$K_2$). \iffinal Frames~$X_1$, $X_2$, $Y_1$,
$Y_2$, $Z_1$ \&~$Z_2$ \copyright~Muchalogy\perm.\fi}}\label{fig:mucha}
\end{figure*}

\begin{figure}[ht]
\def\svgwidth{\hsize}\import{figs/frames/}{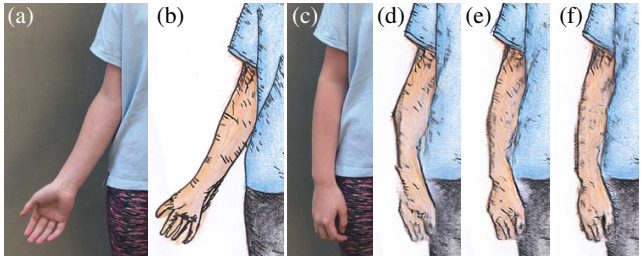}\caption{A different sampling strategy for a selection of frames
in~$Z$---a source frame from a sequence~$V$~(a) and its stylized counterpart~(b)
are used as~$K$. Then weights of~$\F$ are optimized with~$K$ and~$Z$,
where~$Z$ contains all frames from~$V$~(d), $10\%$ of uniformly sampled frames
from~$V$~(e), and~$10\%$ of adaptively sampled frames from~$V$~(f). Note how
dense sampling tends to produce distortion artifacts on a rare hand pose~(c)
due to overfitting on a different pose that is more frequent in the
sequence~$V$~(a) whereas sparse sampling generalizes better.
\iffinal\revision{Source video frames~(a, c) and style~(b)
\copyright~\v{S}t\v{e}p\'{a}nka S\'{y}korov\'{a}\perm.\fi}}\label{fig:frames}
\end{figure}

Video stylization results together with a side-by-side comparison of the output
from previous techniques~\cite{Jamriska19, Texler20b} is presented in
Figures~\ref{fig:teaser} and~\ref{fig:video} as well as in our supplementary
video. In each experiment, we selected a keyframe~$X$ from the input video
sequence~$V$ which was stylized by an artist to produce~$Y$. Then a~$10\%$ of
video frames from~$V$ were sampled uniformly to get the set~$Z$. Using this
input, the weights~$\theta$ of the network~$\F$ were optimized and used to
stylize the entire sequence~$V$. \revision{In~\fg{mucha} we compare the
scenario where multiple keyframes~$K$ are used to stylize~$V$.} We also
considered an option that all frames from~$V$ are used as~$Z$, or instead of
using uniform sampling we selected~$10\%$ of frames that represent the most
signficant changes in the scene. We found that sparse uniform sampling has
usually the best performace~(see~\fg{frames}).

As visible from the results and comparisons, our approach can better preserve
style details during a longer time frame even if the scene structure changes
considerably with respect to~$X$. Also, note how the resulting stylized
sequence has better temporal stability implicitly without performing any
additional treatment, which contrasts with previous techniques~\cite{Jamriska19, Texler20b}
that need to handle temporal consistency explicitly.

\begin{figure*}[ht]
\def\svgwidth{\hsize}\import{figs/boy/}{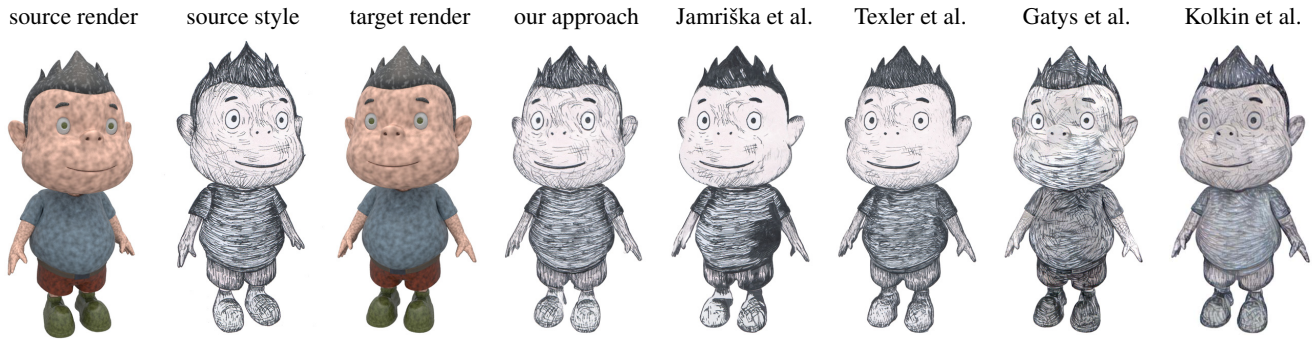}\caption{Stylization of 3D renders---a colored 3D model enhanced with an
artificial noisy texture to avoid large flat regions~(source render) is
stylized at a selected viewpoint by an artist~(source style). The network is
then trained using the stylized pair and a set of additional renders of the
same model viewed from a different direction~(target render). The trained
network can then be used to stylize the rendered 3D model from a different
user-specified position in real-time~(our approach). When compared to other
concurrent style transfer techniques~(Jamri\v{s}ka et al.~\cite{Jamriska19},
Texler et al.~\cite{Texler20b}, Gatys et al.~\cite{Gatys16}, and Kolkin et
al.~\cite{Kolkin19}) our approach better preserves important high-frequency
details of the original style exemplar while being able to adopt to a new
pose in a semantically meaningful way. \iffinal\revision{Source
style~\copyright~\v{S}t\v{e}p\'{a}nka S\'{y}korov\'{a}\perm.\fi}}\label{fig:boy}
\end{figure*}

\begin{figure*}[ht]
\def\svgwidth{\hsize}\import{figs/3d/}{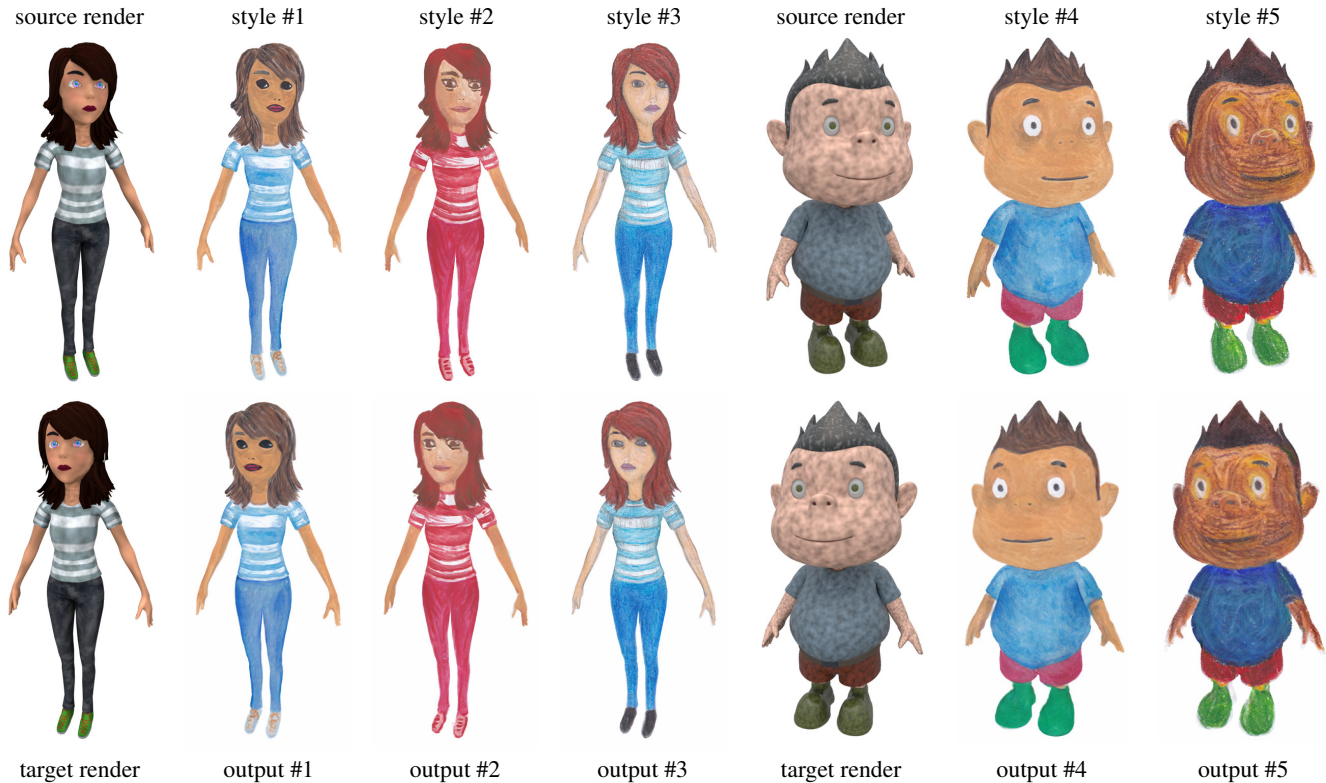}\caption{Stylization of 3D renders (cont.)---a colored 3D model enhanced
by a noisy texture~(source render) is stylized by hand using various artistic
media~(style \#1--\#5). The resulting image translation network~$\F$ is then
used to stylize the same 3D model~(output \#1--\#5) rendered from a different
viewpoint~(target render) in real-time. \iffinal\revision{Source styles
(\#1--\#5) \copyright~\v{S}t\v{e}p\'{a}nka S\'{y}korov\'{a}\perm.}\fi}\label{fig:3d}
\end{figure*}

Style transfer to 3D models resembles video stylization use case, however,
there are specific features worth separate discussion. In this scenario we let
the user select a camera viewpoint from which a 3D model is rendered to get
image~$X$. As the network~$\F$ is sensitive to local variations in~$X$, it is
important to avoid larger flat regions which can make the translation
ambiguous. Due to this reason we add a noisy texture to the 3D model to
alleviate the ambiguity (see source render in~\fg{boy}). An artist then
prepares the stylized counterpart~$Y$ and the model is rendered again from a
few different viewpoints to produce~$Z$. Using those inputs, weights~$\theta$
of the network~$\F$ are optimized and the translation network can then be used
in an interactive scenario where the user changes the camera viewpoint, the 3D
model is rendered on the fly, and immediately stylized using~$\F$. See
Figures~\ref{fig:boy} and~\ref{fig:3d} and our supplementary video for results
in this scenario. As in the video stylization case when compared to other
techniques~\cite{Gatys16, Kolkin19, Jamriska19, Texler20b} our approach better
preserves the style exemplar (c.f.~\fg{boy}) and implicitly maintains temporal
consistency.

\begin{figure*}[ht]
\def\svgwidth{\hsize}\import{figs/panorama/}{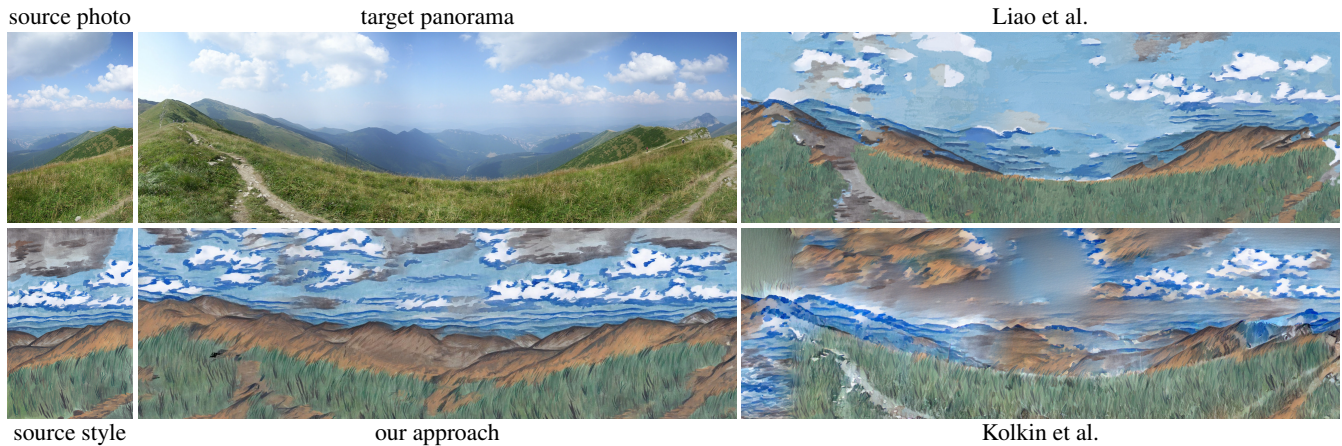}\caption{Panorama stylization results---a photo~(source photo)
is selected from a set of shots taken around the same location by rotating a
camera~(target panorama) and stylized using different artistic media~(source
style). The network is then trained using the stylized pair and a subset of
photos of the panoramic image (target panorama). Finally, the network is used
to stylize each shot, and the entire panorama is stitched together~(our
approach). In contrast to previous techniques (Liao et al.~\cite{Liao17} and
Kolkin et al.~\cite{Kolkin19}) our approach better preserves essential artistic
features and transfers them into appropriate semantically meaningful locations.
See also results with additional styles in~\fg{panos}. \iffinal\revision{Source
style \copyright~\v{S}t\v{e}p\'{a}nka S\'{y}korov\'{a}\perm.}\fi}\label{fig:panorama}
\end{figure*}

\begin{figure}[ht]
\def\svgwidth{\hsize}\import{figs/panorama/}{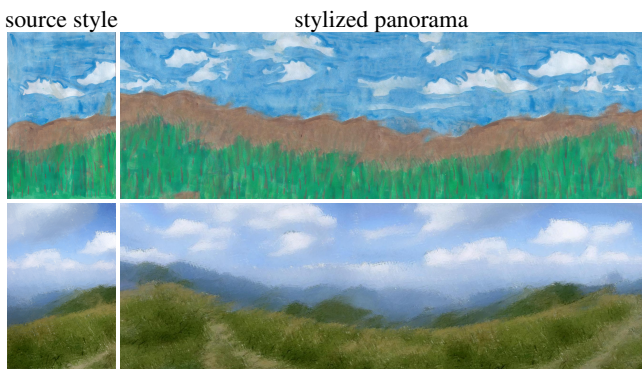}\caption{Panorama stylization results (cont.)---two
additional artistic styles~(source style) used to stylize the panorama shown
in~\fg{panorama}. Note how our approach (stylized panorama) handles also a
higher level of abstraction (first row). \iffinal\revision{Source style~(top
row) \copyright~Jolana S\'{y}korov\'{a}\perm.}\fi}\label{fig:panos}
\end{figure}

In the panorama auto-painting scenario we consider a set of photos~$P$ taken
from the same location by rotating the camera around its center of projection.
We compute a set of homographies~$H$ between photos in~$P$ using the method of
Brown et al.~\cite{Brown07}. Then we let the artist pick one photo from~$P$
as~$X$ and produce its stylized counterpart~$Y$. Remaining photos in~$P$ are
used as~$Z$. After the optimization one can use~$\F$ to stylize all photos
in~$P$, stitch them together using~$H$, and either produce a cylindrical unwrap
or alternatively use an interactive scenario where the user changes the
relative camera rotation from which a pinhole projection can be computed and
stylized in real-time using~$\F$. As visible in~\fg{panorama}
and~\ref{fig:panos} from the comparisons with~\cite{Liao17, Kolkin19} our
approach better preserves the original style details as well as semantic
context.

\begin{figure*}[ht]
\def\svgwidth{\hsize}\import{figs/face/}{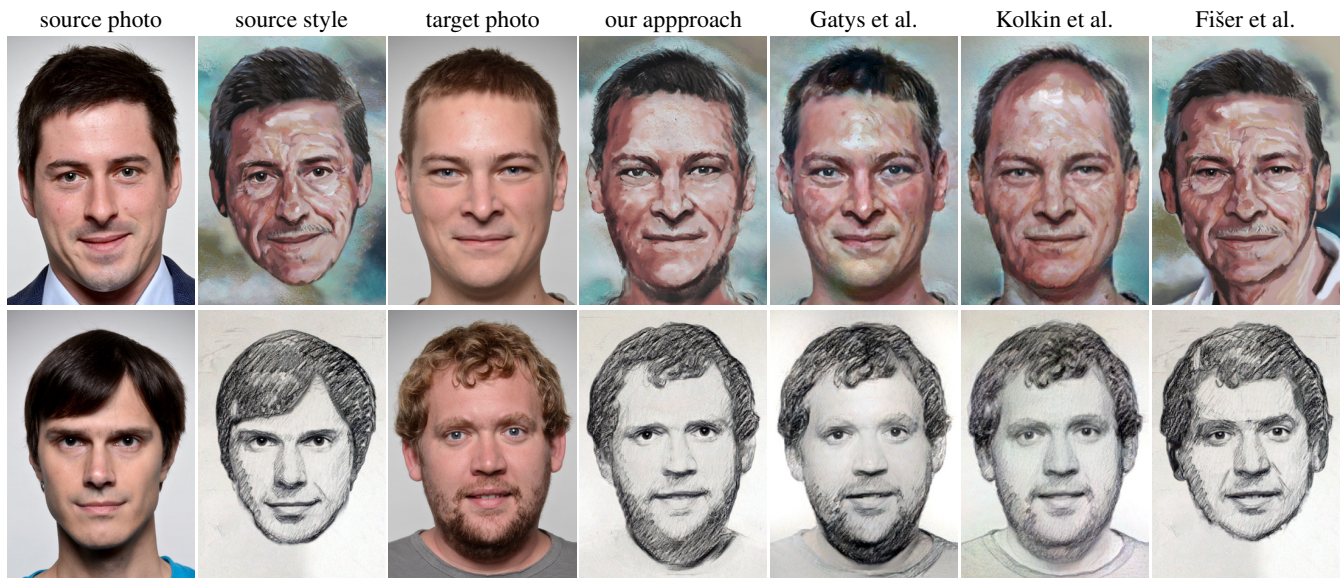}\caption{Stylization of portraits---a portrait photo~(source photo)
taken from a set of portraits captured under similar lighting conditions is
stylized by an artist~(source style). The network is then trained on the
stylized pair and other portraits from the original set~(target photo). Once
trained the network can be used to stylize the other portraits~(our approach).
Even in this more challenging scenario our method produces a reasonable
compromise between style and identity preservation whereas concurrent
techniques suffer either from loosing important high-frequency details~(Gatys
et al.~\cite{Gatys16} and Kolkin et al.~\cite{Kolkin19}) or have difficulties
to retain identity~(Fi\v{s}er et al.~\cite{Fiser17}). \iffinal\revision{Source
style~(top row) \copyright~Graciela Bombalova-Bogra and style~(bottom row)
\copyright~Adrian Morgan\perm.}\fi}\label{fig:face}
\end{figure*}

\def\U{\revision{U}}

In the example-based portrait stylization use case a set of portraits~$\U$ is
assumed to be taken under similar lighting conditions. One portrait from~$\U$
is used as~$X$ and stylized to get~$Y$. The rest of portraits in~$\U$ is used
in~$Z$. Resulting model~$\F$ can then be used to stylize all portraits in~$\U$.
In~\fg{face} stylization results for two different style exemplars are
presented. It is apparent that our approach produces a reasonable compromise
between identity and style preservation whereas previous neural methods such
as~\cite{Gatys16, Kolkin19} tend to preserve identity better, but lose style
details. On the other hand, patch-based technique~\cite{Fiser17} reproduces
style better, nevertheless, has difficulties retaining identity.

\begin{figure}[h]
\def\svgwidth{\hsize}\import{figs/call/}{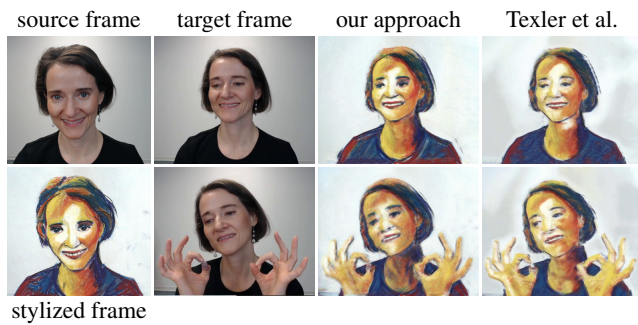}\caption{Real-time stylization of video calls---a frame from a training
sequence~(source frame) is stylized by an artist~(source style). The network
weights are then optimized using this stylized pair and remaining frames from
the training sequence. The final image translation model can be used for
real-time stylization of a new video conference call that contains the same
person and have similar lightihg conditions~(target frames). Note that in
contrast to the method of Texler et al.~\cite{Texler20b} our approach better
preserves style details and keeps the stylization more consistent in time (see
also our supplementary video). \iffinal\revision{Video frames and source style
\copyright~Zuzana Studen\'{a}\perm.}\fi}\label{fig:call}
\end{figure}

In real-time stylization of video calls we let the user record a short video
sequence~$V$ which captures her face during a regular video meet. A most
representative frame is selected from~$V$ and used as~$X$. An artist then
produces its stylized counterpart~$Y$ and~$10\%$ of other frames in~$V$ are
used as~$Z$. A model~$\F$ is optimized using those inputs. Then, during the
next video call~$\F$ is used to stylize captured video frames in real-time.
See~\fg{call} and our supplementary video for an example of such interactive
stylized video call. From the comparison with the method of Texler et
al.~\cite{Texler20b} it is visible that our approach not only better preserves
the overall style quality but also retains temporal stability which is
difficult to accomplish by the method of Texler et al.~in this kind of
interactive scenario.

\revision{\subsection{Perceptual study}}

\revision{In order to qualitatively evaluate our approach, we performed a
perception study comparing the outputs of our method with the outputs of three
state-of-the-art techniques (Jamri\v{s}ka et al.~\cite{Jamriska19}, Kolkin et
al.~\cite{Kolkin19}, and Texler et al.~\cite{Texler20}~(green points)). In our
experiment we wanted to evaluate how well our method reproduces the given
artistic style and how well it preserves the content of the target image. To
perform the evaluation, we collected data via an online survey, where we
presented 170 participants with a randomized set of comparisons (2AFC) asking
to choose which anonymized stylization reproduces style or preserves content
better. In total each participant responded to 28 questions. In each question,
an output from a different method was paired with the output from our technique
using the same input data.}

\revision{The measured preference scores of our method compared to other
techniques can be seen in~\fg{userstudy}. We set out a null hypothesis that
"there is no statistically significant difference in the content preservation
or style reproduction between the results of our method and the other methods."
Then we discussed the probability of rejection of the null hypothesis using the
data we collected via Student's t-test. In the style reproduction category, we
were able to reject the null hypothesis with more then 99\% probability in
comparison to all tested methods in favor of our method. In the content
preservation category, we were able to reject the null hypothesis with more
than 99\% probability, but only the comparison with the method of Jamri\v{s}ka
et al.~was in favor of our method while the other two were not.}

\begin{figure}[ht]
\def\svgwidth{\hsize}\import{figs/userstudy/}{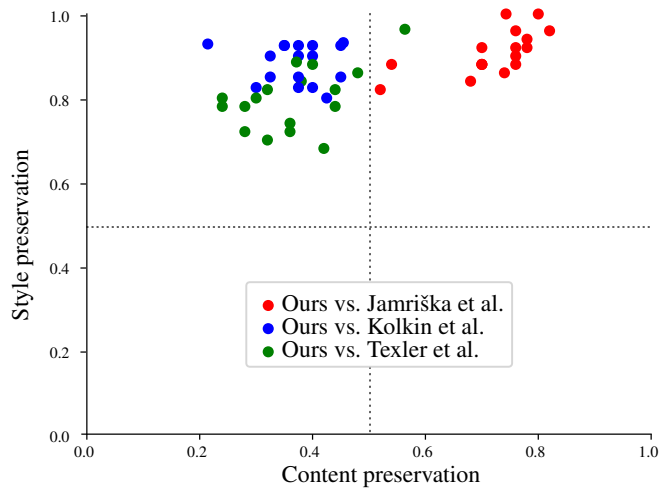}\caption{\revision{Results of perceptual study---each
point represents aggregated votes over a group of 10~participants. On the
$x$~axis we depict the percentage of answers in favor of content preservation
of our method while on the $y$~axis we show the style reproduction percentage.
Comparisons were performed with the method of Jamri\v{s}ka et
al.~\cite{Jamriska19}~(red points), Kolkin et al.~\cite{Kolkin19}~(blue points),
and Texler et al.~\cite{Texler20}~(green points). From the graph it is visible
that our method is observed to reproduce style notably better than previous
works. It also outperforms the method of Jamri\v{s}ka et al.~w.r.t.~the content
preservation, however, Kolkin et al.~as well as Texler et al.~are better in
content preservation.}}\label{fig:userstudy}
\end{figure}

\section{Limitations and Future Work}

While our approach improves on current state-of-the-art in example-based
stylization, we have observed some limitations in how it can be applied.

The most important limitation as compared to related approaches is notably
longer time frame required to finish the optimization, which might be
prohibitive for artist's exploration. To alleviate this drawback we
envision a combination of fast patch-based training strategy of Texler et
al.~\cite{Texler20b} with the computation of VGG loss which needs to be
performed in a full-frame setting.

Due to the usage of relatively computationally expensive neural network model,
the maximum resolution is limited. While we are able to generate output images
with resolutions greater than method of Texler et al.~(e.g.~$768\times 768$
vs.~$512\times 512$), it is still significantly lower than what patch-based
methods~\cite{Jamriska19} are capable of. As a future work we envision to
alleviate this drawback by combining our neural approach with patch-based
technique of~\cite{Texler20}.

In our proposed workflow an artist is responsible for keyframe selection. While
some rules of thumb can be applied, such as selecting a frame that contains all
features that are descriptive for most other frames, a mechanism which would
select the keyframe automatically would improve ease of use.

\begin{figure}[ht]
\def\svgwidth{\hsize}\import{figs/limits/}{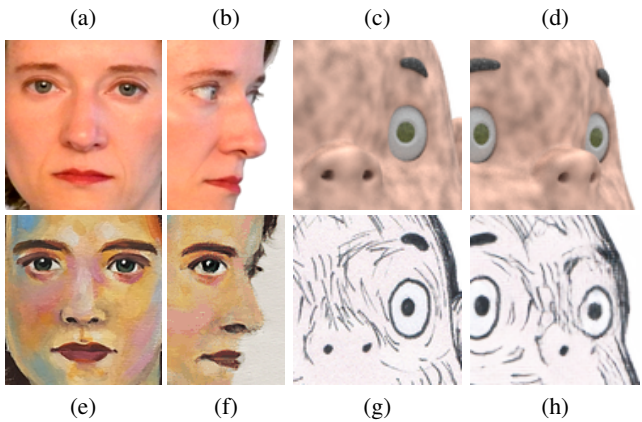}\caption{Limitation on generalization---although our approach
usually generalizes better than concurrent stylization
techniques~\cite{Jamriska19,Texler20b}, some specific features like eyes~(a, c)
that tend to generate strong activation in selected layers of VGG network may
bias the VGG loss and make the network~$\F$ reproduce their mostly unchanged
copies~(f, h) instead of adapting to their actual geometric distortion~(b, d).
\iffinal\revision{Video frames~(a, b) and style~(e) \copyright~Zuzana
Studen\'{a}\perm.}\fi}\label{fig:limits}
\end{figure}

\begin{figure}[ht]
\def\svgwidth{\hsize}\import{figs/zuzka2/}{zuzka2_s.tex}\caption{\revision{The advantage of using style
transfer with auxiliary pairing in visual attribute transfer scenario of Deep
Image Analogy~\cite{Liao17}. Although the style's texture and semantics (see
source style in~\fg{teaser}) are preserved well in both techniques, Deep Image
Analogy~(Liao et al.) has difficulties in adapting to certain structural
changes. \iffinal Target video frame~\copyright~Zuzana Studen\'{a}\perm.\fi}}\label{fig:cmp_liao}
\end{figure}

A key advantage of our approach over current state-of-the-art in example-based
video stylization~\cite{Jamriska19,Texler20b} is greater robustness to
structural discrepancies in the target frames. Even a relatively significant
change such as head rotation is handled relatively well (see~\fg{teaser}). In
this case the network can successfully reproduce newly appearing content while
still being able to preserve the notion of important planar structures of the
original artistic media. On the other hand, some specific localized features
such as eyes, may remain unchanged (see~\fg{limits}). \revision{A similar issue
is known from visual attribute transfer approaches such as Deep Image
Analogy~\cite{Liao17}. As compared to them our method is able to adopt to
structural changes better (see~\fg{cmp_liao}).}

\begin{figure}[ht]
\def\svgwidth{\hsize}\import{figs/ffhq/}{ffhq.tex}\caption{Limitation on a greater appearance change in the target
photo---a key assumption of our method is that the domain of source and target
photos is similar, e.g., photos are taken under same illumination conditions.
When this requirement is not satisfied, the resulting stylization may start to
show artifacts as is visible in those examples of photos taken from the FFHQ
dataset~\cite{Karras19} where the illumination conditions are different to
those used for the capture of source photo in~\fg{face}.}\label{fig:ffhq}
\end{figure}

Most significantly, the method does not seem to generalise very well for
completely generic use cases, for example in~\fg{ffhq}, where input images are
sampled from different underlying distributions. Thus the set of potential
applications is limited to groups of images of visually similar settings
created under comparable conditions.

\section{Conclusion}

We presented an approach of semantically meaningful style transfer that can
leverage a limited number of paired exemplars to stylize a broader set of
target images having similar content to the examples. We optimize weights of an
existing image-to-image translation network by minimizing a novel kind of
objective function that considers the consistency among the provided stylized
pairs as well as discrepancy between VGG features of style exemplars and a
subset of stylized target images.

Thanks to this combination, our approach can
better preserve style details even when the target images' content differs
significantly from the style exemplar. Moreover, our method implicitly
maintains temporal consistency in the video stylization scenario, which needs
to be treated explicitly in previous techniques. We demonstrated the benefits
of our approach in numerous practical use cases, including style transfer to
videos and faces, auto-painting of panorama images, and real-time stylization of
3D models and video calls.

\iffinal

\section*{Acknowledgements}

\revision{We thank the anonymous reviewers for their valuable feedback and
insightful comments. We are also grateful to Zuzana Studen\'{a},
\v{S}t\v{e}p\'{a}nka S\'{y}korov\'{a}, Jolana S\'{y}korov\'{a}, Graciela
Bombalova-Bogra, Adrian Morgan, and Muchalogy for providing style exemplars and
input video sequences. This research was supported by Adobe, the Grant Agency
of the Czech Technical University in Prague, grant No.~SGS19/179/OHK3/3T/13
(Research of Modern Computer Graphics Methods), and by the Research Center for
Informatics, grant No.~CZ.02.1.01/0.0/0.0/16\_019/0000765.}

\fi

\bibliographystyle{eg-alpha-doi}
\bibliography{main}

\end{document}